\documentclass[10pt,twocolumn,letterpaper]{article}

\usepackage{cvpr}              

\usepackage{multirow,mathtools,MnSymbol}
\usepackage{amssymb}
\usepackage{amsthm}
\usepackage{graphicx}
\usepackage{color}
\usepackage[accsupp]{axessibility} 
\usepackage[dvipsnames]{xcolor}

\definecolor{cvprblue}{rgb}{0.21,0.49,0.74}
\usepackage[pagebackref,breaklinks,colorlinks,citecolor=cvprblue]{hyperref}

\def\paperID{1930} 
\def\confName{CVPR}
\def\confYear{2024}

\title{Distributionally Generative Augmentation for Fair Facial Attribute Classification}

\author{
    Fengda Zhang$^{1}\thanks{These authors contributed equally to this work.}$,
    Qianpei He$^{1}\footnotemark[1]$,
    Kun Kuang$^{1}\thanks{Corresponding author.}$,
    Jiashuo Liu$^2$,
    \\
    Long Chen$^3$,
    Chao Wu$^1$,
    Jun Xiao$^1$,
    Hanwang Zhang$^{4,5}$
    \\
    $^1$Zhejiang University \;\; 
    $^2$Tsinghua University \;\;
    $^3$HKUST \;\;
    $^4$NTU \;\;
    $^5$Skywork AI\\
 {\tt\small\{fdzhang,hqp,kunkuang,chao.wu,junx\}@zju.edu.cn} ~
 {\tt\small liujiashuo77@gmail.com} \\
 {\tt\small longchen@ust.hk} ~
 {\tt\small hanwangzhang@ntu.edu.sg}
}

\begin{document}
\maketitle
\begin{abstract}
Facial Attribute Classification (FAC) holds substantial promise in widespread applications. However, FAC models trained by traditional methodologies can be unfair by exhibiting accuracy inconsistencies across varied data subpopulations. This unfairness is largely attributed to bias in data, where some spurious attributes (e.g., Male) statistically correlate with the target attribute (e.g., Smiling). Most of existing fairness-aware methods rely on the labels of spurious attributes, which may be unavailable in practice.
This work proposes a novel, generation-based two-stage framework to train a fair FAC model on biased data without additional annotation. 
Initially, we identify the potential spurious attributes based on generative models.
Notably, it enhances interpretability by explicitly showing the spurious attributes in image space.
Following this, for each image, we first edit the spurious attributes with a random degree sampled from a uniform distribution, while keeping target attribute unchanged. Then we train a fair FAC model by fostering model invariance to these augmentation. 
Extensive experiments on three common datasets demonstrate the effectiveness of our method in promoting fairness in FAC without compromising accuracy.
Codes are in \url{https://github.com/heqianpei/DiGA}.
\end{abstract}    
\section{Introduction}
\label{sec:intro}




\begin{figure}
\begin{center}
\includegraphics[width=\linewidth]{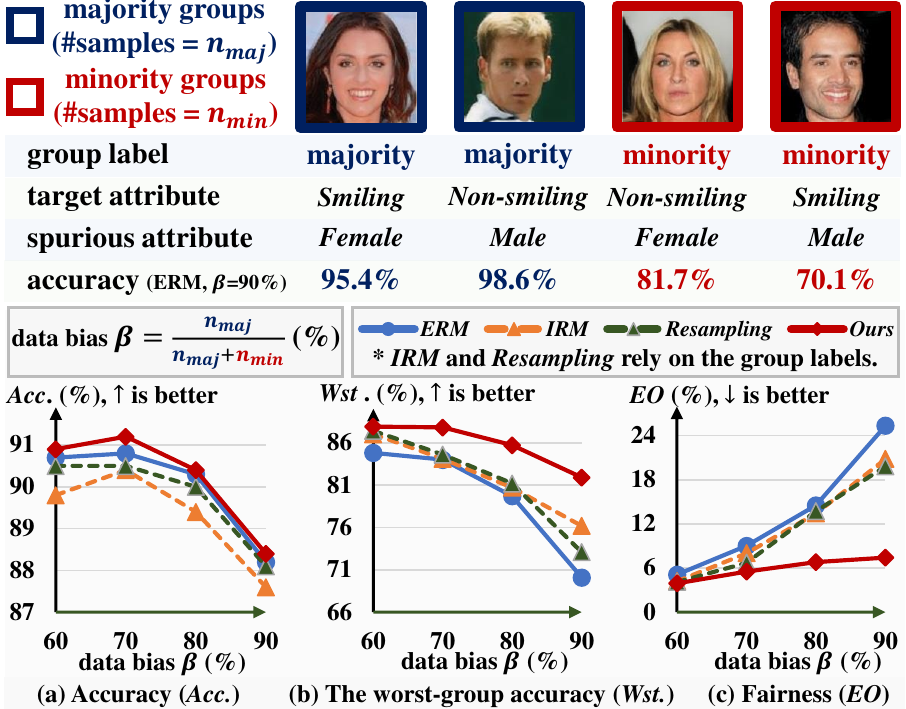}
\end{center}
\vspace{-1.3em}
\caption{FAC models can be unfair by exhibiting accuracy inconsistencies across varied data subpopulations (\textit{e.g.}, 95.4\% accuracy on \textit{Smiling}\&\textit{Female} and  70.1\% accuracy on \textit{Smiling}\&\textit{Male}).
This unfairness is predominantly attributed to data bias, measured by $\beta$.
In general, the more biased the data (\textit{i.e.}, larger $\beta$), the more unfair the model.
Most of existing methods such as IRM~\cite{arjovsky2019invariant} and resampling~\cite{romano2020achieving} rely on the labels of spurious attributes. Our method can improve the fairness, measured by \textit{EO} and the worst-group accuracy (Eq.~(\ref{EO}) and (\ref{worst})), of FAC models without additional annotations. Experiments above are performed on CelebA~\cite{liu2018large}.}
\label{fig-1}
\vspace{-0.4em}
\end{figure}

Facial Attribute Classification (FAC) has garnered significant interest owing to its broad and practical applications like face verification and image retrieval~\cite{zhuang2018multi,thom2020facial}.
The goal of FAC is to predict a certain \textit{target attribute} (\textit{e.g.}, \textit{Smiling}) of a given facial image.
Unfortunately, previous studies have shown that the FAC models can be unfair by exhibiting accuracy inconsistencies across different data subpopulations~\cite{park2022fair}.
This unfairness is predominantly attributed to \textit{bias} in the training data~\cite{mehrabi2021survey,wang2020towards, du2020fairness,gustafson2023facet}.
For example, as shown in Figure \ref{fig-1}, the majority of \textit{Smiling} images in the training dataset are \textit{Female} (termed as \textit{spurious attribute}). Then, the FAC models trained by traditional methods (\textit{e.g.}, Empirical Risk Minimization (ERM)) may use the spurious attribute as a shortcut to predict the target attribute. As a results, the models may suffer from low accuracy on certain data subpopulations (\textit{e.g.}, \textit{Non-smiling}\&\textit{Female}), which seriously hinders their applications in the real world~\cite{liu2019fair}.

To train a fair model on the biased dataset, a number of approaches have been proposed. These methods can be broadly divided into two categories.
The first category mitigates bias by adding a fairness-aware regularization into the training optimization objective~\cite{kamishima2011fairness,donini2018empirical,sagawa2019distributionally}.
However, recent findings indicate that the regularization terms of these methods easily suffer from overfitting, causing such methods to degenerate into ERM~\cite{hu2018does,zhou2022sparse}.
The second category transforms the data by reweighting or augmentation to mitigate bias~\cite{sagawa2020investigation,zhou2022model,kim2023fair}.
Among them, a series of recent studies have achieved great success at fair FAC by using generative models to construct an unbiased dataset~\cite{ramaswamy2021fair,peychev2022latent,li2023augmentation,zhang2023fairnessaware}.


Although existing methods have made strides in improving fairness, most of them require the labels of spurious attributes. Unfortunately, spurious attributes annotation may be unavailable in practice for some reasons~\cite{hashimoto2018fairness,lahoti2020fairness,ashurst2023fairness,choi2020fair}.
Firstly, the vast spectrum of attributes present in images creates a challenge in identifying which are spurious attributes. Secondly, some attributes (\textit{e.g.}, \textit{Attractive}) are difficult to label due to their inherent subjectivity and ambiguity. Lastly, the annotation costs are expensive, especially for large-scale datasets. These considerations naturally prompt the inquiry: \textit{How can one construct a fair FAC model on biased data without the labels of spurious attributes?}

\begin{figure}
\begin{center}
\includegraphics[width=\linewidth]{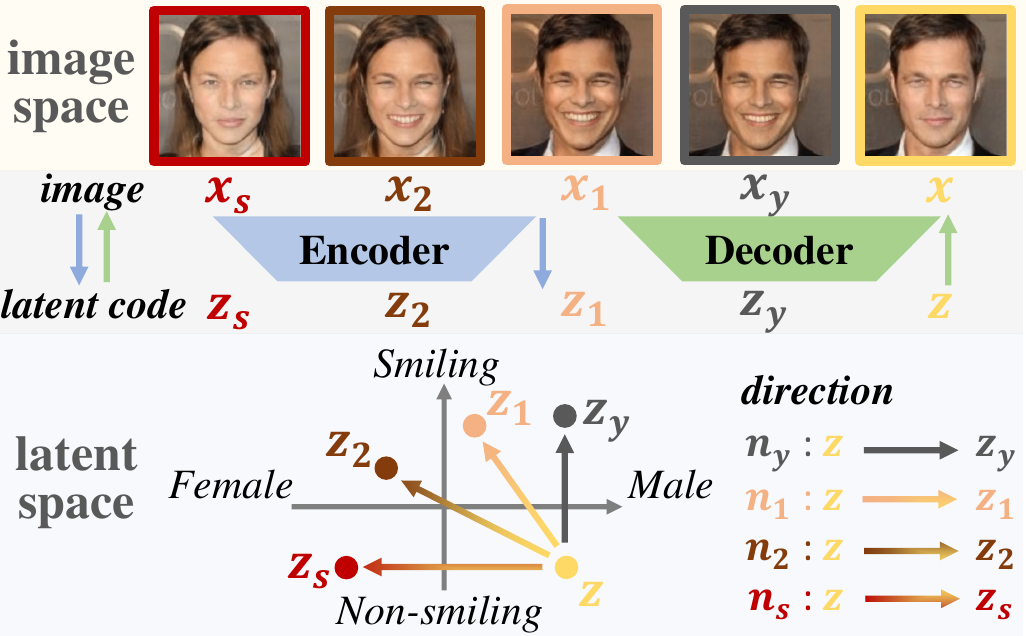}
\end{center}
\vspace{-1.25em}
\caption{Moving the \textit{latent codes} $\boldsymbol{z}$ of a well-trained generative model in a \textit{learned direction} $\boldsymbol{n}$ can edit the target attribute (\textit{e.g.}, \textit{Smiling}) of images ($\boldsymbol{x}\rightarrow \boldsymbol{x}_y$)~\cite{shen2020interfacegan}. We observe that if the data is biased, the learned direction will be \textit{biased} by containing information of spurious attributes (\textit{e.g.}, \textit{Male}) ($\boldsymbol{x}\rightarrow \boldsymbol{x}_1,\boldsymbol{x}_2$). Based on this, we synthesize a direction $\boldsymbol{n}_s$ to manipulate the spurious attributes while keeping target attribute unchanged ($\boldsymbol{x}\rightarrow \boldsymbol{x}_s$).}
\label{fig-2}
\vspace{-1em}
\end{figure}

In this paper, we solve this problem based on a finding in the generative models, as show in Figure~\ref{fig-2}.
Previous studies have shown that \textit{latent codes} for well-trained generative models actually encode disentangled representations, and moving the latent codes $\boldsymbol{z}$ in a \textit{learnable direction} $\boldsymbol{n}$ can manipulate the target attribute (\textit{e.g.}, \textit{Smiling}) of images ($\boldsymbol{x}\rightarrow \boldsymbol{x}_y$)~\cite{shen2020interfacegan,wang2021HFGI}. Based on this, we further observe that the learned direction will be \textit{biased} if there is a spurious correlation in training data. By `biased', we mean that the direction also contains semantics of spurious attributes (\textit{e.g.}, \textit{Male}), so that the potential spurious attributes will change along with the target attribute during editing ($\boldsymbol{x}\rightarrow \boldsymbol{x}_1,\boldsymbol{x}_2$).

Inspired by this finding, we introduce a two-stage framework to address the posed question.
(1) In the first stage, we identify the spurious attributes via generative models. Specifically, we combine two different biased directions (\textit{i.e.}, $\boldsymbol{n}_1$ and $\boldsymbol{n}_2$ in Figure~\ref{fig-2}) in a proper way to cancel out the semantic of target attribute, so that the combined direction $\boldsymbol{n}_{s}$ will only encode spurious attributes. Then, by editing one or more images with the combined direction, the changes of spurious attributes will be faithfully reflected in the image space, as shown in Figure~\ref{fig-2} ($\boldsymbol{x}\rightarrow \boldsymbol{x}_s$).
(2) In the second stage, we learn a fair FAC model via generative augmentation. For each image, we first edit its spurious attributes with a random degree sampled from a \textit{uniform distribution}. Then we train a fair FAC model by promoting its invariance to such augmentations. We call the proposed approach \textit{\textbf{Di}stributionally \textbf{G}enerative \textbf{A}ugmentation} (\textit{DiGA}).

Our method presents two primary merits. Firstly, no annotation outside of target attribute is required. Leveraging generative models, our method mirrors potential data biases within the image space explicitly, concurrently enhancing interpretability. Secondly, the random degree for fairness-aware generative augmentation follows \textit{uniform distribution}. Compared to the existing \textit{single point} augmentations (\textit{e.g.}, flipping the spurious attributes)~\cite{ramaswamy2021fair,zhang2023fairnessaware}, it provides more information for the subsequent fair representation learning and thus enhances the representation quality.

We carried out experiments on CelebA~\cite{liu2018large} and UTKFace~\cite{zhang2017age} datasets for FAC. The classification results in terms of accuracy and fairness show the effectiveness of the proposed \textit{DiGA}. Additionally, we performed extensive analysis experiments to further illustrate the merits of our method in many ways. Moreover, through empirical studies on the Dogs and Cats dataset~\cite{kim2019learning}, we showcased the potential of our approach in general bias mitigation.

Our main contributions can be summaried as: (1) The formulation of an interpretable bias detection technique using generative methods for FAC. (2) The introduction of a fair representation learning strategy predicated on distributionally generative augmentation. (3) Comprehensive experiments across three prevalent datasets, demonstrating that our framework effectively enhances fairness without sacrificing accuracy relative to compared baselines in FAC.

\section{Related work}
\label{sec:related_work}

\noindent\textbf{Bias Mitigation with Group Information.}
There are two main branches to train a fair model on biased data.
The first branch introduces regularizations into optimization objective~\cite{kamishima2011fairness,donini2018empirical,sagawa2019distributionally,levy2020large,ma2023invariant,hanel2022enhancing,xu2021consistent}.
For example, distributionally robust optimization (DRO) methods optimize the worst-case performance~\cite{duchi2018learning}, while invariant risk minimization (IRM) learns unbiased representations with invariance to different environments~\cite{arjovsky2019invariant}.
However, regularization-based methods have proved to be prone to the overpessimism or overfitting problem~\cite{zhou2022model,zhou2022sparse}.
The second branch is to construct an unbiased dataset by transforming data~\cite{sagawa2020investigation,yang2022enhancing}.
Typically, reweighting-based methods reweight the data distribution by some heuristics and train models on the reweighted distribution~\cite{kim2023fair,lv2023duet,lv2024intelligent}. However, the reweighted distribution still has the same support with the original biased distribution. In order to better improves fairness, recent studies have successfully transformed the training distribution by using generative models to generate training samples for minority groups~\cite{ramaswamy2021fair,peychev2022latent,li2023augmentation,zhang2023fairnessaware}.

\noindent\textbf{Bias Mitigation without Group Information.}
Some studies have explored how to mitigate bias without additional annotation~\cite{hashimoto2018fairness,ashurst2023fairness,zhu2023weak,choi2020fair,liu2021heterogeneous,liu2021kernelized,chai2022fairness,yan2020fair,zhao2022towards,romanov2019s}. Most of these methods predict the bias information as the proxies for the spurious attributes by some heuristics (usually the prediction errors given by a biased classifier)~\cite{lahoti2020fairness,nam2020learning,creager2021environment,liu2021just}.
Recently, researchers have tried to improve robustness based on pre-trained models (e.g., CLIP~\cite{radford2021learning}) without additional annotations~\cite{fairclip1, fairclip2, fairclip3, fairclip4, fairclip5, fairclip6}. Note that these efforts rely on pre-trained models, and our work can be done without using pre-trained models. For example, we can use a reference model trained by JTT~\cite{liu2021just} instead of CLIP.

\noindent\textbf{Generative Modeling for Fairness.}
Generative models have achieved great success in recent years~\cite{liu2019stgan,he2019attgan, dogan2020semi,shen2020interfacegan,wang2021HFGI}.
Some works have proposed to evaluate fairness by generating counterfactual samples~\cite{denton2019image, joo2020gender, dash2022evaluating}.
Recently, generation-based methods have demonstrated significant strides in bias mitigation by constructing a balanced and unbiased dataset~\cite{ramaswamy2021fair,peychev2022latent,li2023augmentation,zhang2023fairnessaware,hwang2020fairfacegan}. 
However, these methods need the prior of additional annotations.
In this paper, we extend the generation-based approach to cases without additional annotations.

\noindent\textbf{Fair Representation Learning.}
Learning representations is important for reliable performance in visual recognition. Recent years, contrastive learning has been remarkably successful in learning effective representations~\citep{wang2017community, xiao2020should, zhang2020federated, chen2020simple, he2020momentum, grill2020bootstrap,park2020readme,chen2021exploring}.
However, traditional representation learning methods ignore potential fairness issues. To this end, as a pre-processing method, fair representation learning has achieved great success~\cite{zemel2013learning,madras2018learning,creager2019flexibly,zhao2019inherent,wang2020towards,shen2021contrastive,park2021learning,du2021fairness}. For example, \textit{FSCL} proposes to learn fair representations by closing the distance of samples with the same target attribute labels but different sensitive attribute labels~\citep{park2022fair}. However, most of existing fair representation learning methods rely on labels of spurious attributes, while our method avoids this limitation by the proposed bias detection method. 

\section{Method}
\label{sec:method}

\begin{figure*}
\begin{center}
\includegraphics[width=\linewidth]{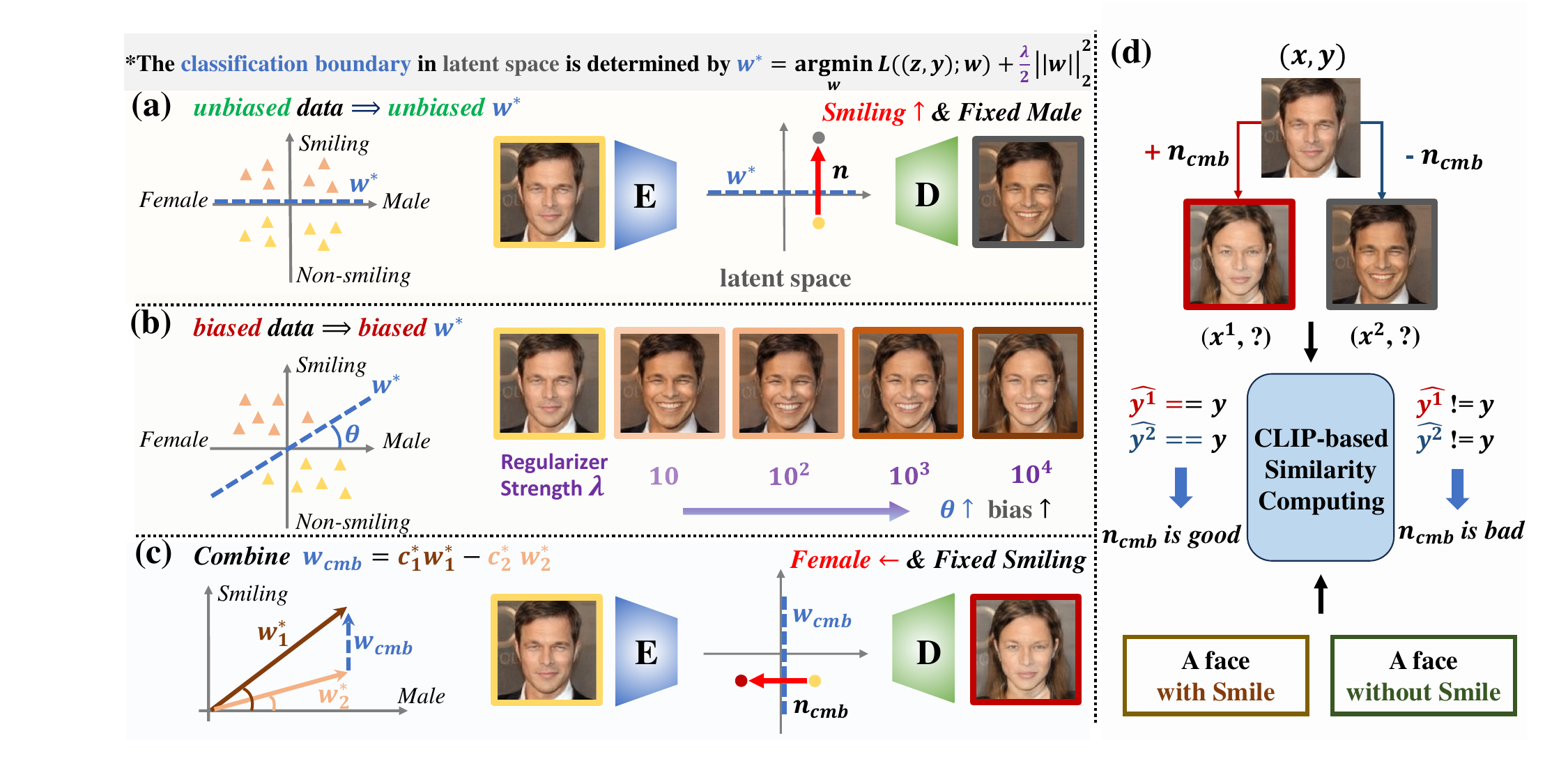}
\end{center}
\vspace{-1.5em}
\caption{\textbf{Illustration of our bias detection method.} (a) Latent codes for well-trained generative models encode disentangled representations, and moving the latent codes along the normal vector $\boldsymbol{n}$ of the learned classification boundary $\boldsymbol{w}^*$ can edit the target attribute of images. (b) The learned boundary $\boldsymbol{w}^*$ will be biased if the training data is biased, and the bias degree of boundary $\boldsymbol{w}^*$ is influenced by the regularization strength $\lambda$. (c) By choosing the appropriate coefficients $(c^*_1,c^*_2)$, we can combine two biased boundaries $(\boldsymbol{w}^*_1,\boldsymbol{w}^*_2)$ into a new boundary $\boldsymbol{w}_{cmb}$ that is only dependent of the spurious attributes. So the direction $\boldsymbol{n}_{cmb}$, the normal vector of $\boldsymbol{w}_{cmb}$, only encode the semantics of spurious attributes. (d) To find the optimal coefficients $(c^*_1,c^*_2)$, we perform grid search with the help of a reference model.}
\label{fig-3}
\vspace{-0.5em}
\end{figure*}

In this section, we introduce our two-stage framework to train a fair FAC model on biased data without the labels of spurious attributes.
We first state our findings in generative modeling on biased data.
Then we propose a generation-based approach for bias detection with theoretical justification.
Finally, we develop a method based on distributionally generative augmentation for fair representation learning.


\subsection{Findings in Biased Generative Modeling}
\textbf{Image Attribute Manipulation via Latent Space.}
Previous works have shown that we can manipulate an image's target attribute (\textit{e.g.}, \textit{Smiling}) via latent space of generative models~\cite{shen2020interfacegan,wang2021HFGI}.
Typically, given a well-trained GAN model, the generator $G:\mathcal{Z} \rightarrow \mathcal{X}$ can map a latent code $\boldsymbol{z}\in\mathcal{Z}$ to an image $\boldsymbol{x}\in\mathcal{X}$, where $\mathcal{Z}$ denotes the latent space.
As shown in Figure~\ref{fig-3}(a), we can train a linear classifier in latent space to learn the boundary hyperplane, with a unit normal vector $\boldsymbol{n}$, of the target attribute. Then, the target attribute of image $\boldsymbol{x}$ can be manipulated by altering its latent code $\boldsymbol{z}$ along the normal vector $\boldsymbol{n}$, \textit{i.e.}, $\boldsymbol{x}_{edit}=G(\boldsymbol{z}\pm\alpha\boldsymbol{n})$, where $\alpha\in\mathbb{R}^+$ controls the degree of image attribute editing.

\noindent\textbf{Biased Semantic Direction.} Consider that some sensitive attributes (\textit{e.g.}, \textit{Male}) have a statistically association with the target attribute in the training data, as shown in Figure~\ref{fig-3}(b). In this case, if we train a linear classifier of target attribute in latent space by regularized logistic regression, the learned classifier (\textit{i.e.}, boundary hyperplane) also will be biased. Therefore, when the latent code of the image is moved along the normal of the biased classification hyperplane, not only the target attribute but also the spurious attributes are changed. Moreover, we also note that the biased degree can be affected by the regularization strength.

\subsection{Bias Detection via Generative Modeling}
\noindent\textbf{Semantic Direction Combination.} In order to detect the potential spurious attributes, our idea is to synthesize a direction of spurious attributes in latent space, so that we can manipulate the spurious attributes while keeping the target attribute unchanged in image space.
To achieve this, we first train a generative model on the training dataset (or a subset, for efficiency).
Then, by using different regularization strengths, we can obtain two different biased semantic directions of target attribute.
Finally, as shown in Figure~\ref{fig-3}(c), we combine these two biased directions by some appropriate combination coefficients. By `appropriate', we mean that the semantic of the target attribute can be cancelled out to zero while only the semantics of spurious attributes are remained. Theoretical guarantees for the existence of optimal combination coefficients are stated later (Theorem~\ref{existence}).
Note that the proposed method naturally supports multi-spurious attribute setting. To extend the method to the multi-class setting, we can transform the problem into multiple binary classification in bias detection stage.

\noindent\textbf{Grid Search for Optimal Combination Coefficients.} The remaining question is how to find the optimal combination coefficients. In this paper, we use the grid search method to traverse the parameter values and then apply the different resultant directions to a small image subset of the training data.
In order to judge whether the components of the target attribute semantics in the direction are cancelled out, we can utilize the artificial judgment or a reference model, as shown in Figure~\ref{fig-3}(d). More implementation details of grid search for coefficients selection are stated in Experiments.

\noindent\textbf{Remark.} We would like to emphasize that previous work has theoretically demonstrated that a model cannot be guaranteed to be fair with only target attribute labels~\cite{lin2022zin}, suggesting that introducing additional information is necessary. Note that the cost of the additional information we used is very low. For artificial judgment, only several images needs to be manually judged. As for the reference model, it is not necessary to have high accuracy. There are many open source vision-language foundation models available such as CLIP~\cite{radford2021learning} that can be used to evaluate the changes to the target attributes after editing by zero-shot prediction. Actually, the assumption of reference model is commonly used in fairness studies~\cite{choi2020fair} since it is easy to obtain in practice.

\subsection{Theoretical Justification for Bias Detection}
Here we provide a theoretical justification for the above bias detection method in a common setting~\cite{sagawa2020investigation}.
Without loss of generality, we start from the setup that both target attribute $y\in\{1,-1\}$ and spurious attribute $s\in\{1,-1\}$ are binary.
Consider that the training dataset of size $n$ is divided into four groups: two majority groups with $s=y$, each containing $n_{maj}/2$ samples, and two minority groups with $s=-y$, each containing $n_{min}/2$ samples. We define the bias degree of data as $\beta=n_{maj}/(n_{maj}+n_{min})\in[1/2,1)$. The larger the $\beta$, the stronger the correlation between $s$ and $y$ in the training data. We say the data is unbiased if $\beta=1/2$. Given a well-trained GAN model, each group has its own distribution over latent codes $\boldsymbol{z}=[\boldsymbol{z}_{y},\boldsymbol{z}_{s}]\in\mathbb{R}^{2d}$ consisting of stable features $\boldsymbol{z}_{y}\in\mathbb{R}^{d}$ generated from the target attribute $y$, and spurious features $\boldsymbol{z}_{s}\in\mathbb{R}^{d}$ generated from the spurious attribute $s$:
\begin{align}
  \boldsymbol{z}_{y}~|~y \sim N(y\textbf{1},\sigma_{y}^{2}I_{d}),
  \\
  \boldsymbol{z}_{s}~|~s \sim N(s\textbf{1},\sigma_{s}^{2}I_{d}).
\end{align}
To get the classification boundary hyperplane, we use regularized logistic regression with optimization objective:
\begin{equation}  
\label{logreg}\mathop{\min}_{\boldsymbol{w}\in\mathbb{R}^{2d}}\mathbb{E}_{(\boldsymbol{z},y)}[\rm{log}(1+exp(-y\boldsymbol{w}\boldsymbol{z}))]+\frac{\lambda}{2}||\boldsymbol{w}||^2_2,
\end{equation}
where $\boldsymbol{w}=[\boldsymbol{w}_y,\boldsymbol{w}_s]$ are linear classifier parameters and $\lambda>0$ controls regularization strength. The parameters of learned classifier are denotes as $\boldsymbol{w}^*=[\boldsymbol{w}^*_y,\boldsymbol{w}^*_s]$, and we define the bias degree of the classifier as $\beta_{clf}=||\boldsymbol{w}^*_s||/||\boldsymbol{w}^*_y||\in[0,+\infty)$. The larger $\beta_{clf}$ is, the more spurious attribute information the learned classifier uses, and thus the greater the degree of deviation of the classification boundary hyperplane in latent space. The classifier is unbiased if and only if $\beta_{clf}=0$. Then we have the following theorem:
\newtheorem{theorem}{Theorem}
\begin{theorem}[Optimal combination coefficients' existence]
\label{existence}
The learned classifier with optimization objective \ref{logreg} is biased (\textit{i.e.}, $\beta_{clf}>0$), if the data is biased (\textit{i.e.}, $\beta>1/2$). Moreover, there exists optimal combination coefficients $(c^*_1,c^*_2)\in\mathbb{R}^2_+$ such that $\boldsymbol{w}_{cmb}:=c^*_1\boldsymbol{w}^*_1-c^*_2\boldsymbol{w}^*_2=[\boldsymbol{0},\boldsymbol{1}]$ is dependent of $s$ and independent of $y$, if $\lambda_1 \ll \lambda_2$, where $\boldsymbol{w}^*_1$ and $\boldsymbol{w}^*_2$ are parameters of linear classifiers trained with regularization strengths $\lambda_1$ and $\lambda_2$, respectively.
\end{theorem}
Please refer to Appendix for proof. 
The optimal combination coefficients yield a classifier of potential spurious attributes in latent space, independent of the target attribute.
Theorem \ref{existence} reveals the existence of optimal combination coefficients, laying a foundation for the proposed traversal search-based method for bias detection.

\subsection{Bias Mitigation via Generative Augmentation}
To prevent the model from learning or amplifying potential bias in the training data, we first learn fair representations that contain as little spurious attributes information as possible.
Note that the obtained optimal combined direction $\boldsymbol{n}_{cmb}$ can be used to manipulate the spurious attributes of images while keeping the target attribute unchanged.
Following this, our idea is to train a representation model $E(\cdot, \boldsymbol{\phi})$ with invariance to changes in spurious attributes.

\begin{figure}
\begin{center}
\includegraphics[width=\linewidth]{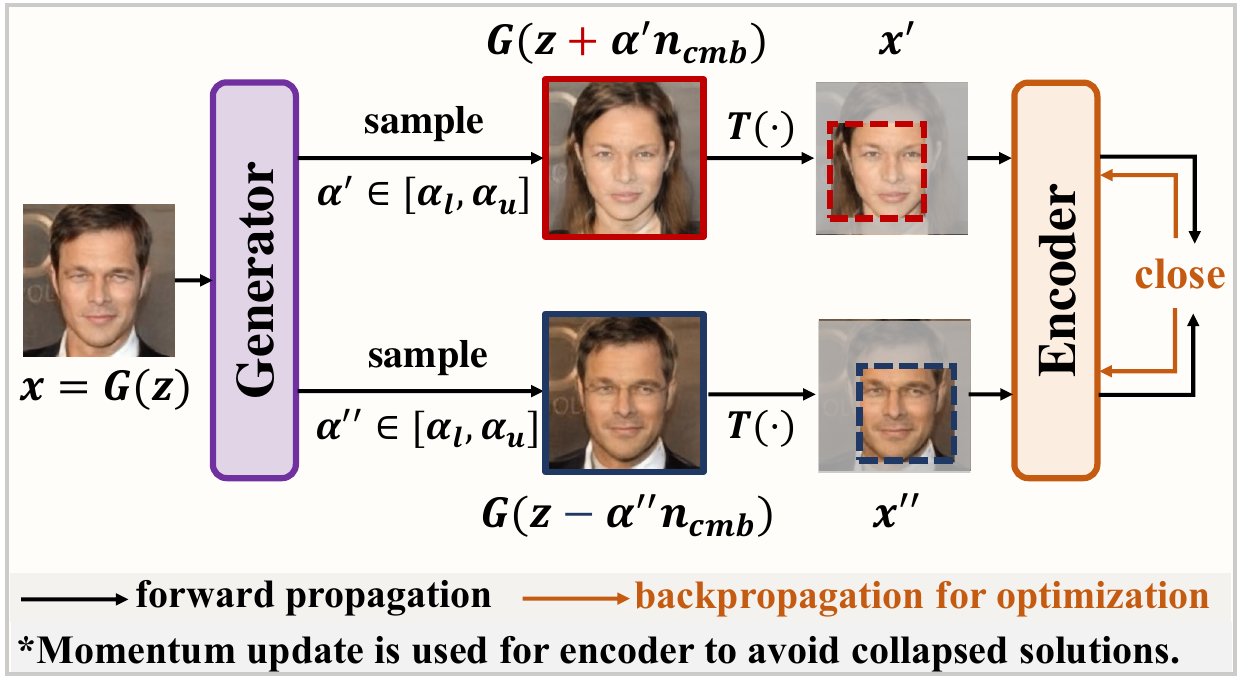}
\end{center}
\vspace{-1.5em}
\caption{\textbf{Distributionally generative augmentation for fair representation learning.} For each image, we edit its spurious attributes by using the combined semantic direction $\boldsymbol{n}_{cmb}$ in latent space. The editing degrees $\alpha'$ and $\alpha''$ are randomly sampled from a uniform distribution. We also perform traditional augmentation $T(\cdot)$ such as random clipping. The encoder is trained to learn fair and effective representations by closing the distance between augmented views. We use momentum encoder to avoid collapsing.}
\label{fig-4}
\vspace{-0.7em}
\end{figure}

We implement this idea based on contrastive learning, as shown in Figure \ref{fig-4}.
Specifically, for each image $\boldsymbol{x}$ in training dataset with latent code $\boldsymbol{z}$, we perform random augmentation, including not only the traditional strategies $T(\cdot)$ such as random clipping, but also the generative augmentation of spurious attributes via direction $\boldsymbol{n}_{cmb}$. By this way, in each iteration, we can get the augmented positive sample pair $\boldsymbol{x}'=T(G(\boldsymbol{z}+\alpha'\boldsymbol{n}_{cmb}))$ and $\boldsymbol{x}''=T(G(\boldsymbol{z}-\alpha''\boldsymbol{n}_{cmb}))$, where $\alpha'$ and $\alpha''$ are uniformly sampled from $[\alpha_l,\alpha_u]$, and $\alpha_l,\alpha_u\in\mathbb{R}^+$ are hyperparameters controlling the variation range of spurious attributes' semantics.
Then, we train a fair encoder $E(\cdot, \boldsymbol{\phi})$ by minimizing the distance between representations of positive samples on training dataset.
Finally, we train a linear classifier $C(\cdot, \boldsymbol{\omega})$ on the top of frozen encoder $E(\cdot, \boldsymbol{\phi})$ on training dataset for fair classification.

Notably, The degree for spurious attributes manipulation follows a uniform distribution rather than a single point.
This distributionally generative augmentation provides finer supervision information as guidance for fair representation learning, thus helping to improve representation quality.
We used techniques such as momentum update and stopping gradient like previous works~\cite{grill2020bootstrap,chen2021exploring} in our implementation.

\section{Experiments}

\begin{figure*}
\begin{center}
\includegraphics[width=6.8in]{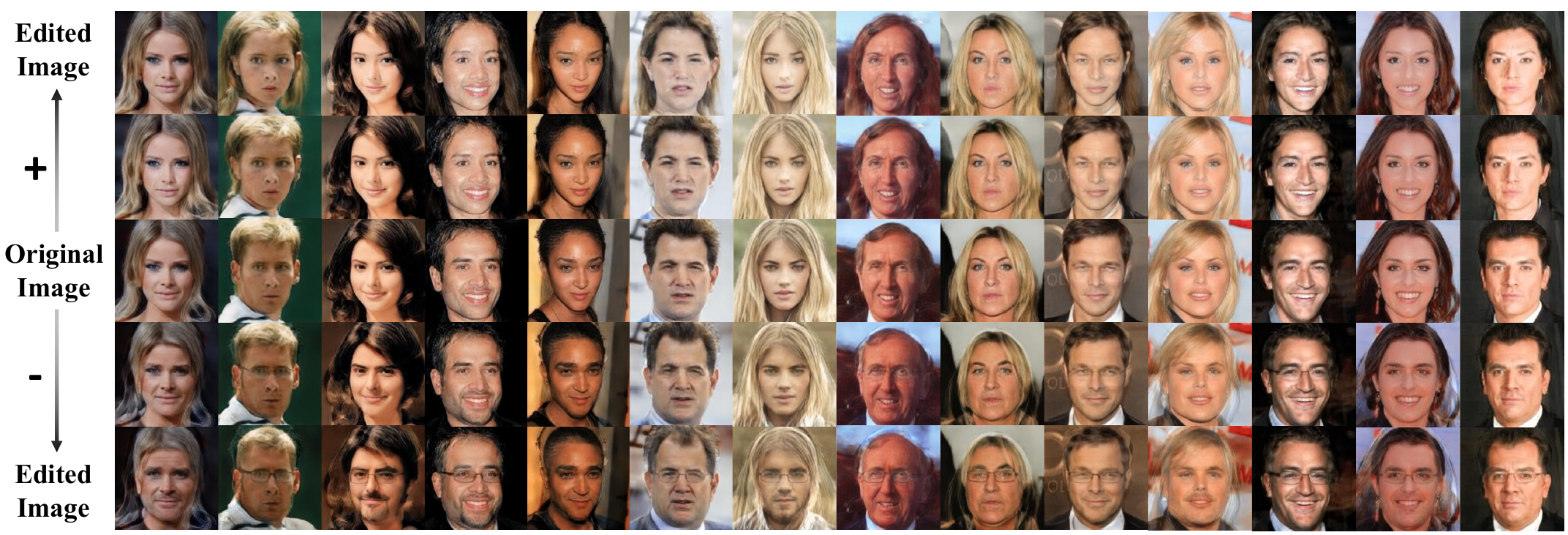}
\end{center}
\vspace{-1.5em}
\caption{\textbf{Bias detection results on CelebA dataset.} The constructed training dataset is biased, where 75\% \textit{Smiling} images are \textit{Female}\&\textit{Young} and 75\% \textit{Non-smiling} images are \textit{Male}\&\textit{Non-young}. We only have labels of target attribute \textit{Smiling}. By utilizing the proposed bias detection method, we obtain the combined direction and edit training images. It can be observed that the changes of gender and age are faithfully reflected in image space, illustrating which attributes are spurious attributes explicitly and thus enhancing interpretability.}
\label{fig-detection}
\vspace{-0.7em}
\end{figure*}

\subsection{Datasets}
To validate our method, we did experiments on identical three datasets that were utilized in prior studies~\cite{ramaswamy2021fair,park2022fair,zhang2023fairnessaware}.

\noindent\textbf{CelebA}~\citep{liu2018large} is a common dataset used for FAC. Each image has 40 binary attributes labels. Following the setting of the previous works~\citep{ramaswamy2021fair,liu2021just,wang2022fairness}, we select \textit{Smiling}, \textit{Blond Hair}, \textit{Black Hair}, \textit{Male} and \textit{Young} as the target attributes, and set \textit{Male} and \textit{Young} as spurious attributes. Besides, to verify the performance of our method in the setting of multi-target labels and multi-spurious attributes, we also set \{\textit{Blond Hair}, \textit{Black Hair}\} as target attributes and \{\textit{Male}, \textit{Young}\} as spurious attributes respectively.
For each experiment, we randomly sample a biased subset as training dataset with size of 20,000 images, where the majority group and minority group have 90\% and 10\% of the sample size respectively. 
We report performance on the whole original test dataset.

\noindent\textbf{UTK-Face}~\citep{zhang2017age} contains over 20k facial images, each with attributes labels. Following the experimental setup in the previous works~\cite{park2022fair,jung2022learning}, we define a binary spurious attribute \textit{Ethnicity} based on whether the facial image is white or not. The task is to predict the \textit{Gender}. We randomly sample a biased subset consisting of 10,000 images, with the same bias degree as CelebA. We also construct a balanced and unbiased test dataset consisting of 3,200 images.

\noindent\textbf{Dogs and Cats} dataset is widely used for general bias mitigation~\cite{kim2019learning,park2022fair,zhang2023fairnessaware}. It contains dog and cat images with additional annotations for partial images about the color of dog/cat is bright or dark. The task is to predict if the image is a cat or a dog and the spurious attribute is color. The biased training set consists of 400 bright cat images, 3,600 bright dog images, 3,600 dark cat images, and 400 dark cat images. The balanced test set consists of 2,400 images.

\subsection{Evaluation Metrics}
Our goal is to learn a fair and accurate FAC model. In this paper, we use equalized odds (\textit{EO})~\citep{hardt2016equality}, one of the most commonly used notion of group fairness~\citep{dwork2012fairness}, as the fairness metric.
Following~\citep{jung2022learning}, we extend \textit{EO} to multi-target attribute and multi-spurious attribute setting:
\begin{equation}
\mathop{\max}_{\forall y, \hat{y}\in \mathcal{Y} \atop \forall s^i, s^j\in \mathcal{S}}\left|P_{s^i}(\hat{Y}=\hat{y} \mid Y=y)-P_{s^j}(\hat{Y}=\hat{y} \mid Y=y)\right|,
\label{EO}
\end{equation}
where $Y$ is ground truth, $\hat{Y}$ is predictive label given by the classifier, and $s^i, s^j \in \mathcal{S}$ is the value of spurious attributes. A smaller \textit{EO} means a fairer classifier.
We also report the worst-group accuracy defined as:
\begin{equation}
\mathop{\min}_{\forall y\in\mathcal{Y} \atop \forall s\in\mathcal{S}} P_{s}(\hat{Y}=y \mid Y=y),
\label{worst}
\end{equation}
Besides, we use accuracy (\%) to measure the model utility.

\subsection{Bias Detection Results on Facial Datasets}
\noindent\textbf{Bias Detection Results.}
Consider that the task is to predict whether a given facial image is \textit{Smiling} or not. The training dataset is constructed to be biased, where the target attribute \textit{Smiling} statistically correlates with two potential spurious attributes \textit{Male} and \textit{Young}. Note that only target attribute labels are available during training.
We use the proposed bias detection method to obtain the combined semantic direction, and edit the training images to detect the bias.
The results on CelebA are shown in Figure \ref{fig-detection}. We can observe that the changes of the potential spurious attributes \textit{Male} and \textit{Young} are faithfully reflected in the image space.

\begin{table}[]
\setlength\tabcolsep{2pt}
\scalebox{0.77}{
\begin{tabular}{cccccccccccc}
\hline
$c_1/c_2$           & 0.5  & 0.6  & 0.7  & 0.8  & \textbf{0.9}  & 1.0  & 1.1  & 1.2  & 1.3  & 1.4  & 1.5  \\
\textit{consistency (\%)} & 73.0 & 76.0 & 78.5 & 83.0 & \textbf{84.5} & 83.5 & 83.0 & 82.5 & 81.0 & 77.5 & 76.5 \\ \hline
\end{tabular}}
\caption{Grid search results for optimal combination coefficients.}
\label{table-search}
\vspace{-1em}
\end{table}

\begin{table*}[]
\setlength\tabcolsep{2.5pt}
\resizebox{\textwidth}{!}{
\begin{tabular}{cccccccccccccccccccccccccccc}
\hline
         & \multicolumn{3}{c}{T=\textit{s} / S=\textit{m}}                & \multicolumn{3}{c}{T=\textit{s} / S=\textit{y}}                & \multicolumn{3}{c}{T=\textit{b} / S=\textit{m}}                 & \multicolumn{3}{c}{T=\textit{a} / S=\textit{y}}                 & \multicolumn{3}{c}{T=\textit{m} / S=\textit{y}}                & \multicolumn{3}{c}{T=\textit{y} / S=\textit{m}}                 & \multicolumn{3}{c}{T=\textit{b}\&\textit{a}\&\textit{r} / S=\textit{m}}           & \multicolumn{3}{c}{T=\textit{s} / S=\textit{m}\&\textit{y}}              & \multicolumn{3}{c}{T=\textit{g} / S=\textit{e}}                \\ \cline{2-28} 
             Method  & \textit{Acc.}           & \textit{Wst.}         & \textit{EO}           & \textit{Acc.}           & \textit{Wst.}         & \textit{EO}           & \textit{Acc.}           & \textit{Wst.}         & \textit{EO}            & \textit{Acc.}           & \textit{Wst.}         & \textit{EO}            & \textit{Acc.}           & \textit{Wst.}         & \textit{EO}           & \textit{Acc.}           & \textit{Wst.}         & \textit{EO}            & \textit{Acc.}           & \textit{Wst.}         & \textit{EO}            & \textit{Acc.}           & \textit{Wst.}         & \textit{EO}            & \textit{Acc.}           & \textit{Wst.}         & \textit{EO}           \\ \hline
\textit{ERM}~\cite{he2016deep}            & 88.2          & 70.1          & 25.3         & 88.3          & 71.5          & 15.6         & 84.2          & 73.3          & 17.1          & 82.8          & 70.1          & 19.4          & 97.2          & 92.8          & 5.4          & 77.7          & 42.0          & 52.0          & 90.6          & 69.3          & 24.1          & 87.3          & 60.4          & 33.8          & 91.4          & 83.5          & 12.2         \\
\textit{CVaR DRO}~\cite{levy2020large}       & 87.3          & 74.0          & 22.8         & 87.0          & 76.1          & 13.9         & 84.0          & 73.9          & 15.5          & 81.4          & 71.8          & 15.2          & 96.5          & 93.0          & 5.3          & 75.4          & 42.3          & 48.8          & 90.0          & 71.8          & 22.0          & 86.3          & 64.0          & 28.4          & 90.6          & 84.5          & 11.9         \\
\textit{EIIL}~\cite{creager2021environment}           & 87.9          & 75.6          & 19.7         & 87.9          & 72.5          & 13.3         & 84.1          & 73.9          & 15.7          & 81.9          & 73.3          & 14.4          & 96.2          & 93.3          & 4.9          & 77.5          & 45.6          & 39.2          & 90.4          & 71.5          & 22.0          & 86.4          & 60.8          & 19.7          & 89.2          & 84.3          & 8.3          \\
\textit{LfF}~\cite{nam2020learning}            & 87.1          & 77.5          & 17.0         & 85.3          & 72.9          & 14.3         & 84.0          & 74.0          & 15.1          & 82.4          & 72.5          & 14.2          & 97.1          & 92.9          & 5.1          & 77.4          & 44.2          & 43.6          & 89.8          & 70.8          & 20.5          & 85.0          & 62.5          & 26.6          & 86.7          & 84.6          & 11.1         \\
\textit{JTT}~\cite{liu2021just}            & 88.0          & 74.8          & 19.4         & 87.6          & 73.3          & 14.2         & 83.9          & 74.1          & 16.7          & 81.1          & 71.1          & 16.6          & 97.0          & 92.4          & 5.8          & 76.3          & 43.6          & 47.7          & 88.3          & 69.1          & 23.3          & 87.3          & 61.0          & 31.0          & 90.5          & 85.0          & 10.4         \\
\textit{MAPLE}~\cite{zhou2022model}          & 88.1          & 72.0          & 19.6         & 88.1          & 73.6          & 13.6         & 83.7          & 73.9          & 14.7          & 82.4          & 74.7          & 13.8          & 97.1          & 92.9          & 4.8          & 76.3          & 46.2          & 43.5          & 89.9          & 72.8          & 18.6          & 86.0          & 64.8          & 31.2          & 89.4          & 85.3          & 9.4          \\
\textit{DiGA} (ours) & \textbf{88.4} & \textbf{81.9} & \textbf{7.4} & \textbf{89.1} & \textbf{78.5} & \textbf{9.5} & \textbf{84.5} & \textbf{74.5} & \textbf{13.5} & \textbf{83.6} & \textbf{78.6} & \textbf{10.8} & \textbf{97.4} & \textbf{94.8} & \textbf{4.3} & \textbf{80.0} & \textbf{51.3} & \textbf{33.3} & \textbf{90.7} & \textbf{79.7} & \textbf{15.8} & \textbf{88.4} & \textbf{75.8} & \textbf{15.6} & \textbf{92.7} & \textbf{89.0} & \textbf{6.8} \\ \hline
\end{tabular}}
\caption{\textbf{Classification results on facial datasets.} We use classification accuracy (\textit{Acc.}), the worst-group accuracy (\textit{Wst.}), and equalized odds (\textit{EO}) to measure the performance of FAC model on CelebA and UTKFace datasets. T and S represent target and spurious attributes, respectively. \textit{s}, \textit{b}, \textit{a}, \textit{r}, \textit{m}, \textit{y}, \textit{g}, and \textit{e} respectively denote \textit{Smiling}, \textit{Blond Hair}, \textit{Black Hair}, \textit{Brown Hair}, \textit{Male}, \textit{Young}, \textit{Gender}, and \textit{Ethnicity}. The spurious attribute labels are unavailable for all methods during training. All the results are the averaged scores over five runs.}
\label{table-cls}
\end{table*}

\begin{figure*}
\begin{minipage}[t]{0.52\textwidth}
\centering
\includegraphics[width=0.95\linewidth]{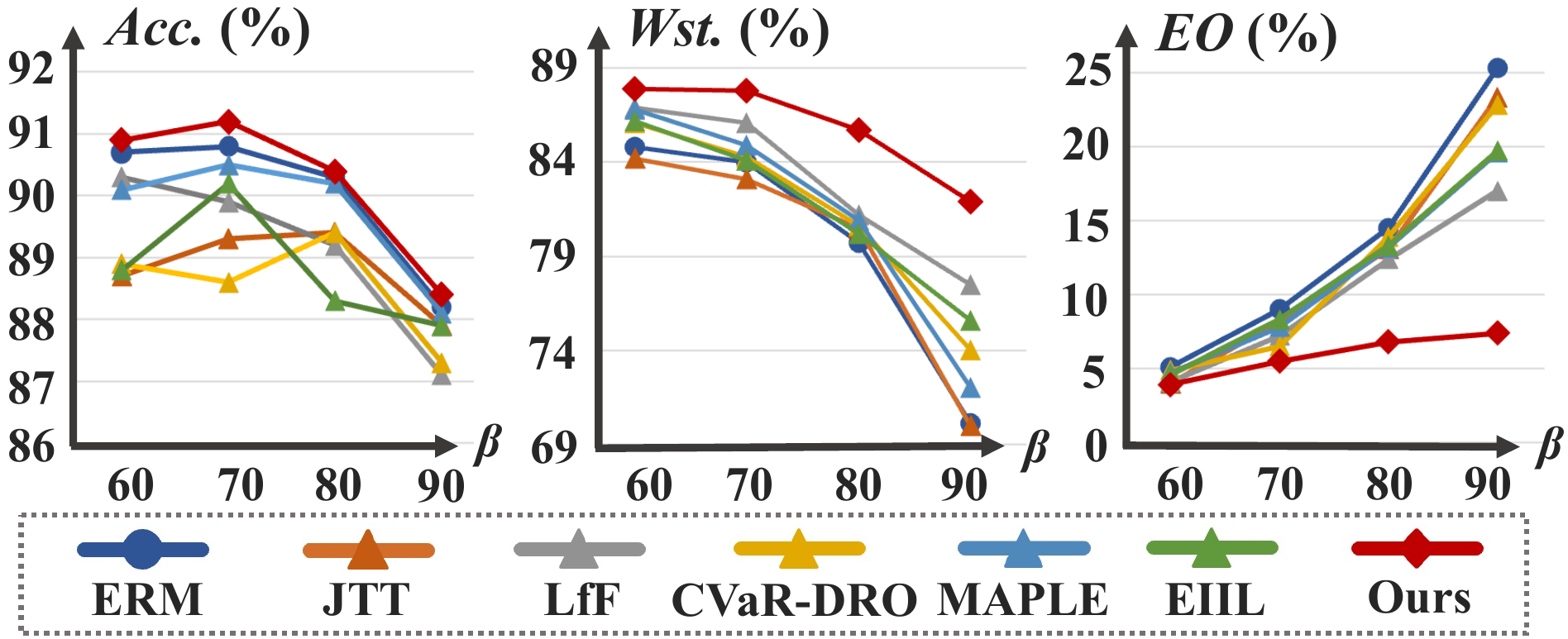}
\caption{\textbf{Classification results on CelebA dataset under different degrees of data bias.} Data bias $\beta\coloneqq\frac{n_{maj}}{n_{maj}+n_{min}}$, where $n_{maj}$ and $n_{min}$ respectively denote the sample size of majority and minority groups.}
\label{fig-bias}
\end{minipage}
\hfill
\begin{minipage}{0.45\textwidth}
    \centering
    \setlength\tabcolsep{2.5pt}
\vspace{-4.5em}
\scalebox{0.85}{
\begin{tabular}{cccccccccc}
\hline
                        & \multicolumn{3}{c}{label ratio=50\%}        & \multicolumn{3}{c}{label ratio=25\%}        & \multicolumn{3}{c}{label ratio=10\%}        \\ \cline{2-10} 
Method                  & \textit{Acc.} & \textit{Wst.} & \textit{EO} & \textit{Acc.} & \textit{Wst.} & \textit{EO} & \textit{Acc.} & \textit{Wst.} & \textit{EO} \\ \hline
\textit{ERM}            & 87.5          & 67.8          & 26.1        & 87.1          & 65.9          & 27.7        & 86.9          & 62.8          & 28.9        \\
\textit{CVaR DRO}       & 86.6          & 72.9          & 22.1        & 86.6          & 72.3          & 22.4        & 85.5          & 69.1          & 27.3        \\
\textit{EIIL}           & 86.2          & 71.3          & 22.5        & 85.9          & 69.6          & 25.4        & 86.8          & 64.2          & 26.7        \\
\textit{LfF}            & 86.9          & 75.5          & 19.4        & 85.9          & 72.1          & 23.6        & 85.5          & 66.1          & 27.7        \\
\textit{JTT}            & 87.3          & 72.9          & 20.1        & 86.7          & 71.1          & 20.6        & 86.8          & 67.1          & 23.1        \\
\textit{MAPLE}          & 87.4          & 73.7          & 23.8        & 87.0            & 72.7          & 24.2        & 85.6          & 69.2          & 27.1        \\
\textit{DiGA} (ours) & \textbf{88.4}          & \textbf{81.1}          & \textbf{7.8}         & \textbf{88.4}          & \textbf{78.3}          & \textbf{8.0}           & \textbf{88.3}          & \textbf{78.8}          & \textbf{8.4}         \\ \hline
\end{tabular}
}
    \captionof{table}{\textbf{Classification results on CelebA dataset} (T=\textit{s}, S=\textit{m}) \textbf{under settings of incomplete target labels.} We set the label ratio of target attribute as 50\%, 25\%, and 10\% respectively.}
    \label{table-incomplete}
  \end{minipage}%
\end{figure*}

\noindent\textbf{Implementations.} 
We perform grid search for the optimal combination coefficients $c^*_1$, $c^*_2$ such that the combined direction $c^*_1\boldsymbol{n}_1-c^*_2\boldsymbol{n}_2$ only contains semantics of spurious attributes.
The criterion is to make the target attribute match most consistent before and after editing, that is, to make the changes of the target attributes as little as possible.
In order to improve efficiency, we randomly sample only 100 images from the training dataset. For each search, we edit them positively and negatively respectively by the combined direction to get 200 edited images. Then we predict their target labels by using CLIP as reference model with prompts ``A face with/without smile".
For ease of search, we rewrite the combined direction as $c_1(\boldsymbol{n}_1-c_2/c_1\boldsymbol{n}_2)$. So we only need to search for $c_2/c_1$ (from 0.5 to 1.5 with unit 0.1). The results are shown in Table \ref{table-search}. We set $c_1/c_2=0.9$, which makes the target attribute change the least after editing.

\subsection{Classification Results on Facial Datasets}
\noindent\textbf{Main Results.}
The classification results on CelebA and UTKFace datasets are shown in Table \ref{table-cls}.
We measure classification accuracy (\textit{Acc.}), the worst-group accuracy (\textit{Wst.}), and \textit{EO} of trained FAC models.
We find that \textit{ERM} achieves good accuracy but suffer from severe unfairness.
We also compare our method with various state-of-the-art debiasing baselines that do not require spurious attribute labels including regularization-based methods (\textit{CVaR DRO}~\cite{levy2020large} and \textit{EIIL}~\cite{creager2021environment}) and reweighting-based methods (\textit{LfF}~\cite{nam2020learning}, \textit{JTT}~\cite{liu2021just}, and \textit{MAPLE}~\cite{zhou2022model}).
We find that although these debiasing methods improve fairness to some extent, they sacrifice accuracy more or less.
Compared with them, our \textit{DiGA} achieves better performance in terms of accuracy, the worst-group accuracy, and \textit{EO} in various settings.

\noindent\textbf{Robustness to Data Bias Degree.}
In Figure \ref{fig-bias}, we show the robustness of different algorithms to the degree of data bias $\beta$ on CelebA (T=\textit{s}, S=\textit{m}), where $\beta$ is defined as the proportion of the majority group sample to the total sample. We can observe that our method achieves state-of-the-art performance in terms of accuracy and fairness under various data bias degrees. Moreover, as the degree of data bias increases, the \textit{EO} gap between the compared baselines and our method gradually increases. This shows that our method has better robustness to the degree of data bias.

\noindent\textbf{Semi-supervised Classification Results.}
Our approach relies only on the target labels without the need for additional annotations. However, target labels are not always available due to annotation costs, which motivates us to test the performance of different methods under the setting of incomplete target labels. We set different label ratios, as shown in Table \ref{table-incomplete}. Compared with baselines, our method has better robustness to ratio of target labels.
The reason may be that our method trains encoder in self-supervised way, while the target labels are only used to train the linear classifiers.

\begin{figure*}[t]
\begin{minipage}[t]{0.48\textwidth}
\centering
\includegraphics[width=3.3in]{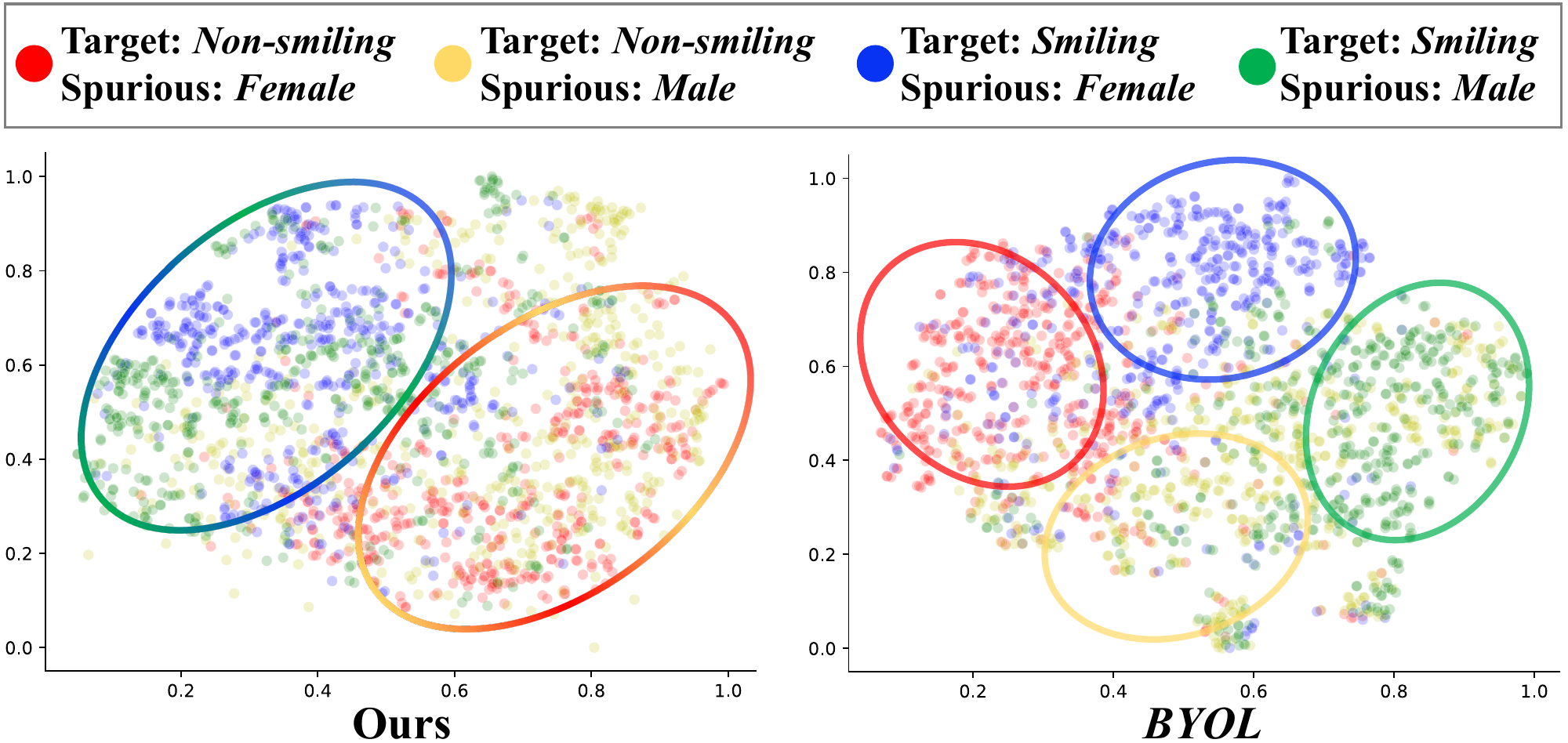}
\caption{\textbf{T-SNE visualization for the learned representations on CelebA} (T=\textit{s}, S=\textit{m}). Compared with BYOL, the representations trained by ours contain less information of spurious attributes.}
\label{fig-tsne}
\end{minipage}
\hfill
\begin{minipage}[t]{0.48\textwidth}
\centering
\includegraphics[width=2.9in]{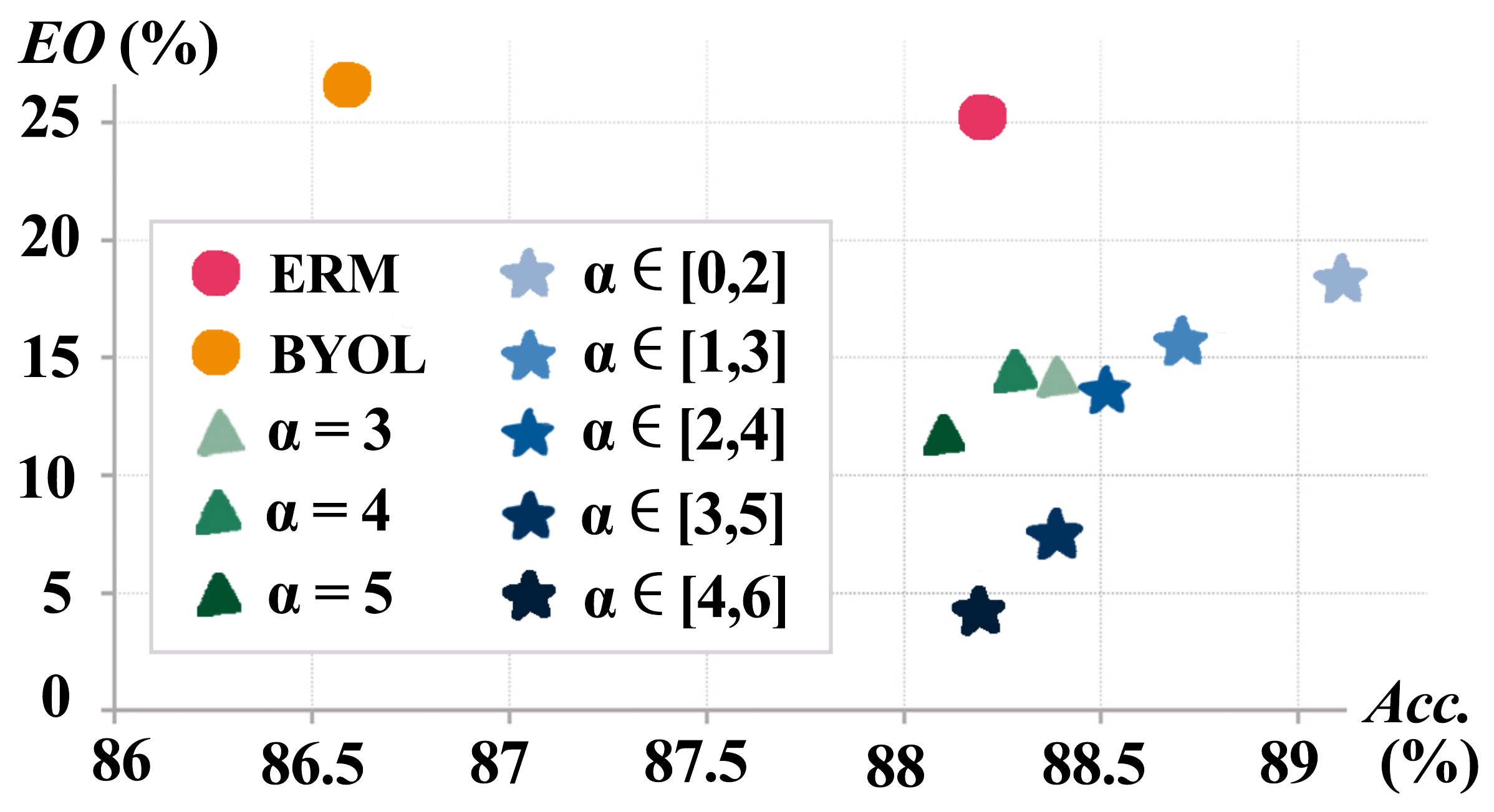}
\caption{\textbf{Ablation studies on generative augmentation.} $\blacktriangle$ and $\bigstar$ respectively denote single point generative augmentation strategy and our distributionally generative augmentation strategy.}
\label{fig-degree}
\end{minipage}
\end{figure*}

\begin{figure*}
\begin{minipage}[t]{0.48\textwidth}
\centering
\includegraphics[width=\linewidth]{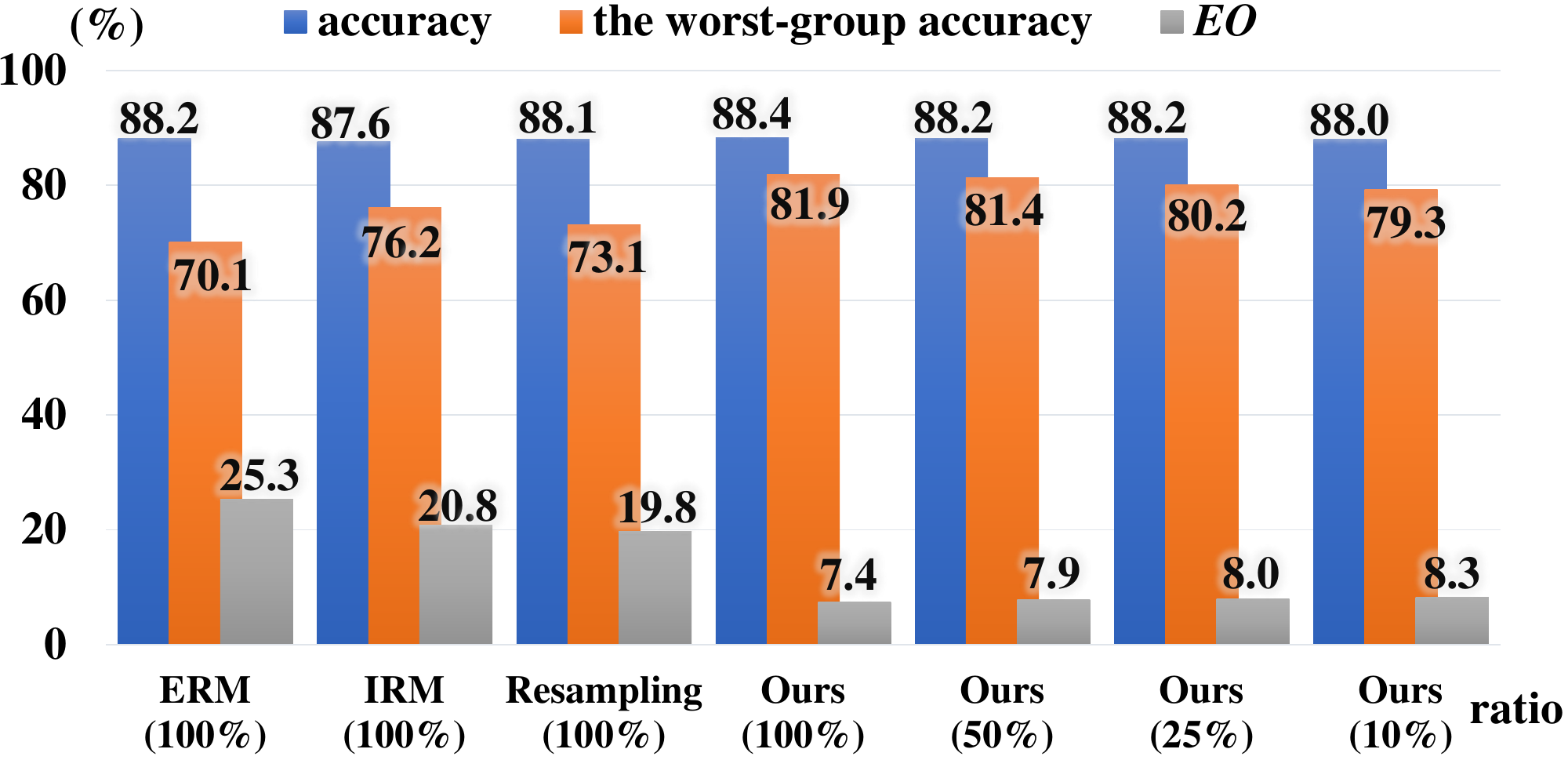}
\caption{\textbf{Empirical studies of sample ratio used for generative modeling.} For efficiency, we can randomly sample a subset to train the generative model, with little final performance degradation.}
\label{fig-efficiency}
\end{minipage}
\hfill
\begin{minipage}{0.23\textwidth}
\vspace{-6.5em}
    \centering
    \setlength\tabcolsep{2.5pt}
\scalebox{0.83}{
\begin{tabular}{clll}
\hline
                 & \multicolumn{1}{c}{\textit{Acc.}} & \multicolumn{1}{c}{\textit{Wst.}} & \multicolumn{1}{c}{\textit{EO}} \\ \hline
$\lambda_1$=2e-4 & \multirow{2}{*}{88.3}             & \multirow{2}{*}{82.7}             & \multirow{2}{*}{9.3}            \\
$\lambda_2$=5e+3 &                                   &                                   &                                 \\ \hline
$\lambda_1$=1e-4 & \multirow{2}{*}{88.4}             & \multirow{2}{*}{81.9}             & \multirow{2}{*}{7.4}            \\
$\lambda_2$=1e+4 &                                   &                                   &                                 \\ \hline
$\lambda_1$=2e-5 & \multirow{2}{*}{88.8}             & \multirow{2}{*}{85.1}             & \multirow{2}{*}{4.8}            \\
$\lambda_2$=5e+4 &                                   &                                   &                                 \\ \hline
$\lambda_1$=1e-6 & \multirow{2}{*}{88.8}             & \multirow{2}{*}{82.3}             & \multirow{2}{*}{7.4}            \\
$\lambda_2$=1e+6 &                                   &                                   &                                 \\ \hline
\end{tabular}
}
\captionof{table}{\textbf{Ablation studies of regularization strength $\lambda_1$, $\lambda_2$ on CelebA} (T=\textit{s}, S=\textit{m}). We set several groups of $\lambda_1$, $\lambda_2$, where $\lambda_1$ is much smaller than $\lambda_2$.}
\label{table-regularization}
  \end{minipage}%
  \hfill
  \begin{minipage}{0.23\textwidth}
  \vspace{-6.5em}
    \centering
    \setlength\tabcolsep{2.5pt}
\scalebox{0.83}{
\begin{tabular}{cccc}
\hline
                        & \multicolumn{3}{c}{T=\textit{s} / S=\textit{c}}               \\ \cline{2-4} 
Method                  & \textit{Acc.} & \textit{Wst.} & \textit{EO} \\ \hline
\textit{ERM}            & 87.5          & 67.8          & 26.1        \\
\textit{CVaR DRO}       & 86.6          & 72.9          & 22.1        \\
\textit{EIIL}           & 86.2          & 71.3          & 22.5        \\
\textit{LfF}            & 86.9          & 75.5          & 19.4        \\
\textit{JTT}            & 87.3          & 72.9          & 20.1        \\
\textit{MAPLE}          & 87.4          & 73.7          & 23.8        \\
\textit{DiGA} (ours) & \textbf{88.4}          & \textbf{81.1}          & \textbf{7.8}         \\ \hline
\end{tabular}
}
\captionof{table}{\textbf{Classification results on non-facial dataset Dogs and Cats.} \textit{s} denotes the target attribute \textit{Species} and \textit{c} denotes the spurious attribute \textit{Color}.}
\label{table-dogsandcats}
  \end{minipage}
\end{figure*}

\subsection{T-SNE Visualization}
To further explain how our method works, we provide visualization of the learned representations via t-SNE~\cite{van2008visualizing} in Figure \ref{fig-tsne}.
We divide the CelebA dataset into four groups in terms of target and spurious attributes and randomly sample 500 images from each group.
We find that traditional representation learning method BYOL~\cite{grill2020bootstrap} learns information of spurious attributes, so that the representations trained by BYOL can be divided by the spurious attributes. In contrast, the representations learned by ours contain less information of spurious attributes, thus contributing to fair classification.

\subsection{Ablation Studies}
\noindent\textbf{Ablation Studies on Generative Augmentation.}
The ablation study results of generative augmentation on CelebA dataset (T=\textit{s}, S=\textit{m}) are shown in Figure \ref{fig-degree}.
Our approach has advantages in the following aspects:
(1) \textit{Comparison with traditional augmentation.}
Compared with the typical contrastive learning BYOL~\cite{grill2020bootstrap} that only performs traditional augmentation (\textit{e.g.}, random cropping), our method achieves better accuracy and fairness thanks to the fairness-aware generative augmentation. 
(2) \textit{Comparison with single point augmentation.} Compared with single point generative augmentation~\cite{zhang2023fairnessaware}, our distributionally generative augmentation achieves better accuracy and fairness. Because it considers a wider data distribution and provides more supervision information for fair representation learning.
(3) \textit{Trade-off between accuracy and fairness.} We can flexibly balance accuracy and fairness via the range of augmentation degree, while larger degree result in a fairer model.

\noindent\textbf{Ablation Studies on Sampling Ratio for Efficient Modeling.}
To improve the efficiency of generative modeling, we implement the sampling strategy. Specifically, we randomly sample a subset to train the generative model and obtain the combined semantic direction. The final classification results on CelebA dataset (T=\textit{s}, S=\textit{m}) are shown in Figure \ref{fig-efficiency}.
We can observe that the performance of our method decreases very little even when the sampling ratio is small (\textit{e.g.}, 10\%).

\noindent\textbf{Ablation Studies on Regularization Strength.}
In Table \ref{table-regularization}, we show the classification results by using different regularization strengths. It shows that our method can achieve good results as long as $\lambda_1$ is much smaller than $\lambda_2$.

\subsection{General Bias Mitigation on Non-facial Dataset}
To verify the effectiveness of our method on visual data other than faces, we perform experiments on Dogs and Cats dataset, where the target and spurious attributes are \textit{Species} and \textit{Color} respectively. The classification results are shown in Table \ref{table-dogsandcats}. Our method achieves better accuracy and fairness than other compared baselines, and it bodes well for the potential of our approach for general bias mitigation. 

\section{Conclusions}
In this paper, we proposed a generation-based two-stage framework to train a fair FAC model on biased data without additional annotations.
In the first stage, we detect the spurious attributes via generative models. Our method enhances interpretability by explicitly representing the spurious attributes in the image space.
In the second stage, for each image, we first edit its spurious attributes, where the editing degree follows a uniform distribution. Then we train a fair FAC model by promoting its invariance to these augmentation.
Extensive experiments on the three datasets demonstrate the effectiveness of our approach.
In future work, we aim to extend our method to support various visual data, with the help of rapidly developing generative models.

~\\
\noindent\textbf{Acknowledgment}
This work was supported by the National Natural Science Foundation of China (62337001, 62376243, 62037001, U20A20387), the Fundamental Research Funds for the Central Universities (No. 226-2022-00051), the StarryNight Science Fund of Zhejiang University Shanghai Institute for Advanced Study (SN-ZJU-SIAS-0010), Project by Shanghai AI Laboratory (P22KS00111) and National Research Foundation, Singapore under its AI Singapore Programme (AISG Award No: AISG2-RP-2021-022). Long Chen is supported by HKUST Special Support for Young Faculty (F0927) and HKUST Sports Science and Technology Research Grant (SSTRG24EG04).
\newpage
{
    \small
    \bibliographystyle{ieeenat_fullname}
    \bibliography{main}
}
\clearpage
\setcounter{page}{1}
\maketitlesupplementary

\noindent The supplementary materials are organized as follows:
\begin{itemize}
\item In Appendix A, we give the proof for Theorem \ref{existence}. Theorem \ref{existence} guarantees the existence of optimal combination coefficients, so that we can use grid search to find them;
\item In Appendix B, as an empirical supplement to Theorem \ref{existence}, we show our observations on synthetic dataset to reveal the relationship between $\beta_{clf}$ (the bias of learned classifier in latent space) and $\lambda$ (the regularization strength); 
\item In Appendix C, we present the additional results of bias detection on real facial dataset to more intuitively show why and how our approach works.
\item In Appendix D, we present the implementation details.
\end{itemize}

\section*{Appendix A. Proof for theoretical justification}
\textit{Proof:}
We first define the sample ratio of majority group and minority group as $p_{maj}=n_{maj}/(n_{maj}+n_{min})$ and $p_{min}=n_{min}/(n_{maj}+n_{min})$ respectively. The optimization objective $R(\boldsymbol{w})$ can be written as
\begin{equation}
\begin{aligned}
R(\boldsymbol{w}) =& \mathbb{E}_{(\boldsymbol{z},y)}[\rm{log}(1+e^{-y\boldsymbol{w}\boldsymbol{z}})]+\frac{\lambda}{2}||\boldsymbol{w}||^2_2
\\
=& \frac{p_{maj}}{2}\mathbb{E}_{\boldsymbol{z}_y\sim N(\textbf{1},\sigma_{y}^{2}I_{d})}\mathbb{E}_{\boldsymbol{z}_s\sim N(\textbf{1},\sigma_{s}^{2}I_{d})}[\rm{log}(1+e^{-\boldsymbol{w}\boldsymbol{z}})]
\\
+& \frac{p_{maj}}{2}\mathbb{E}_{\boldsymbol{z}_y\sim N(\textbf{-1},\sigma_{y}^{2}I_{d})}\mathbb{E}_{\boldsymbol{z}_s\sim N(\textbf{-1},\sigma_{s}^{2}I_{d})}[\rm{log}(1+e^{\boldsymbol{w}\boldsymbol{z}})]
\\
+& \frac{p_{min}}{2}\mathbb{E}_{\boldsymbol{z}_y\sim N(\textbf{1},\sigma_{y}^{2}I_{d})}\mathbb{E}_{\boldsymbol{z}_s\sim N(\textbf{-1},\sigma_{s}^{2}I_{d})}[\rm{log}(1+e^{-\boldsymbol{w}\boldsymbol{z}})]
\\
+& \frac{p_{min}}{2}\mathbb{E}_{\boldsymbol{z}_y\sim N(\textbf{-1},\sigma_{y}^{2}I_{d})}\mathbb{E}_{\boldsymbol{z}_s\sim N(\textbf{1},\sigma_{s}^{2}I_{d})}[\rm{log}(1+e^{\boldsymbol{w}\boldsymbol{z}})]
\\
+& \frac{\lambda}{2}||\boldsymbol{w}||^2_2.
\end{aligned}
\end{equation}
Without loss of generality, we let $d=1$. Then we have
\begin{equation}
\begin{aligned}
R(\boldsymbol{w})
=& \frac{p_{maj}}{2}\mathbb{E}_{z_y\sim N(1,\sigma_{y}^{2}),z_s\sim N(1,\sigma_{s}^{2})}[\rm{log}(1+e^{-w_yz_y-w_sz_s})]
\\
+& \frac{p_{maj}}{2}\mathbb{E}_{z_y\sim N(-1,\sigma_{y}^{2}),z_s\sim N(-1,\sigma_{s}^{2})}[\rm{log}(1+e^{w_yz_y+w_sz_s})]
\\
+& \frac{p_{min}}{2}\mathbb{E}_{z_y\sim N(1,\sigma_{y}^{2}),z_s\sim N(-1,\sigma_{s}^{2})}[\rm{log}(1+e^{-w_yz_y-w_sz_s})]
\\
+& \frac{p_{min}}{2}\mathbb{E}_{z_y\sim N(-1,\sigma_{y}^{2}),z_s\sim N(1,\sigma_{s}^{2})}[\rm{log}(1+e^{w_yz_y+w_sz_s})]
\\
+& \frac{\lambda}{2}||\boldsymbol{w}||^2_2
\\
=& p_{maj}\mathbb{E}_{z_y\sim N(1,\sigma_{y}^{2}),z_s\sim N(1,\sigma_{s}^{2})}[\rm{log}(1+e^{-w_yz_y-w_sz_s})]
\\
+& p_{min}\mathbb{E}_{z_y\sim N(1,\sigma_{y}^{2}),z_s\sim N(1,\sigma_{s}^{2})}[\rm{log}(1+e^{-w_yz_y+w_sz_s})]
\\
+& \frac{\lambda}{2}||\boldsymbol{w}||^2_2.
\end{aligned}
\end{equation}
For convenience, we write $\mathbb{E}_{z_y\sim N(1,\sigma_{y}^{2})}$ and $\mathbb{E}_{z_s\sim N(1,\sigma_{s}^{2})}$ as $\mathbb{E}_{z_y}$ and $\mathbb{E}_{z_s}$ respectively without causing any ambiguity.
Our goal is to minimize $R(w_y,w_s)$. So we focus on the gradients of classifier parameters $w_y$ and $w_s$:
\begin{equation}
\begin{aligned}
&\nabla_{w_y}R(w_y,w_s)
\\
=& p_{maj}\mathbb{E}_{z_y}\mathbb{E}_{z_s}[\frac{1}{(1+e^{w_yz_y+w_sz_s})}(-z_y)]
\\
+& p_{min}\mathbb{E}_{z_y}\mathbb{E}_{z_s}[\frac{1}{(1+e^{w_yz_y-w_sz_s})}(-z_y)]
\\
+&\lambda w_y
\end{aligned}
\end{equation}
and
\begin{equation}
\begin{aligned}
&\nabla_{w_s}R(w_y,w_s)
\\
=& p_{maj}\mathbb{E}_{z_y}\mathbb{E}_{z_s}[\frac{1}{(1+e^{w_yz_y+w_sz_s})}(-z_s)]
\\
+& p_{min}\mathbb{E}_{z_y}\mathbb{E}_{z_s}[\frac{1}{(1+e^{w_yz_y-w_sz_s})}z_s]
\\
+&\lambda w_s.
\end{aligned}
\end{equation}
We use proof by contradiction. Let $w^*_s$ be zero. Then we have
\begin{equation}
\begin{aligned}
&\nabla_{w_s}R(w^*_y,0)
\\
=& p_{maj}\mathbb{E}_{z_y}\mathbb{E}_{z_s}[\frac{1}{(1+e^{w^*_yz_y})}(-z_s)]
\\
+& p_{min}\mathbb{E}_{z_y}\mathbb{E}_{z_s}[\frac{1}{(1+e^{w^*_yz_y})}z_s]
\\
=& (-p_{maj})\mathbb{E}_{z_y}\mathbb{E}_{z_s}[\frac{1}{(1+e^{w^*_yz_y})}z_s]
\\
+& (1-p_{maj})\mathbb{E}_{z_y}\mathbb{E}_{z_s}[\frac{1}{(1+e^{w^*_yz_y})}z_s]
\\
=& (1-2p_{maj})\mathbb{E}_{z_y}\mathbb{E}_{z_s}[\frac{1}{(1+e^{w^*_yz_y})}z_s]
\\
=& (1-2p_{maj})\mathbb{E}_{z_y}[\frac{1}{(1+e^{w^*_yz_y})}]\mathbb{E}_{z_s}[z_s]
\\
=& (1-2p_{maj})\mathbb{E}_{z_y}[\frac{1}{(1+e^{w^*_yz_y})}]
\\
<&0.
\end{aligned}
\end{equation}
Note that $\mathbb{E}_{z_y}[\frac{1}{(1+e^{w^*_yz_y})}]>0$, so that the $\nabla_{w_s}R(w^*_y,0)=0$ if and only if the majority group sample ratio $p_{maj}=1/2$ (\textit{i.e}., the data is unbiased). The above equation shows that the solution $w^*_s$ cannot be zero. Similarly, we also have
\begin{equation}
\nabla_{w_y}R(0,w^*_s)<0.
\end{equation}
So the bias degree of the classifier $\beta_{clf}=||{w}^*_s||/||{w}^*_y||>0$ if the data is biased (\textit{i.e.}, $\beta=p_{maj}=>1/2$). 
Different values of $\lambda$ will scale the impact of the regularization term, affecting the solution $\boldsymbol{w^*}=(w^*_y,w^*_s)$ of logistic regression. Denote the solutions under regularization strength $\lambda_1$ and $\lambda_2$ are $\boldsymbol{w^*_1}=(w^*_{y1},w^*_{s1})$ and $\boldsymbol{w^*_2}=(w^*_{y2},w^*_{s2})$ respectively.
As we have proven before, $w^*_{y1}$, $w^*_{s1}$, $w^*_{y2}$, and $w^*_{s2}$ are not zero.
Then we construct $c^*_1=w^*_{y2}/(w^*_{y2}w^*_{s1}-w^*_{y1}w^*_{s2})$ and $c^*_2=w^*_{y1}/(w^*_{y2}w^*_{s1}-w^*_{y1}w^*_{s2})$ such that $\boldsymbol{w}_{cmb}:=c^*_1\boldsymbol{w}^*_1-c^*_2\boldsymbol{w}^*_2=[0,1]$. Here we have completed the proof of the existence of the optimal combination coefficients. $\qed$



\section*{Appendix B. Observations on synthetic dataset}
\vspace{-0.5em}
In this section, as an empirical supplement to Theorem \ref{existence}, we explore the relationship between $\beta_{clf}$ (bias of learned linear classifier in the latent space) and $\lambda$ (regularization strength used in logistic regression) on synthetic dataset.

\noindent\textbf{Experimental Setup.} Following the previous studies~\cite{sagawa2020investigation}, we use the same settings as in the theoretical justification. Specifically, target attribute $y\in\{1,-1\}$ and spurious attribute $s\in\{1,-1\}$ are binary.
The training dataset contains $n=20000$ samples, which can be divided into four groups: two majority groups with $s=y$, each containing $n_{maj}/2$ samples, and two minority groups with $s=-y$, each containing $n_{min}/2$ samples. In the latent space of generative models, each group has its own distribution over latent codes $\boldsymbol{z}=[\boldsymbol{z}_{y},\boldsymbol{z}_{s}]\in\mathbb{R}^{200}$ consisting of stable features $\boldsymbol{z}_{y}\in\mathbb{R}^{100}$ generated from the target attribute $y$, and spurious features $\boldsymbol{z}_{s}\in\mathbb{R}^{100}$ generated from the spurious attribute $s$: $\boldsymbol{z}_{y}~|~y \sim N(y\textbf{1},\sigma_{y}^{2}I_{100})$ and $\boldsymbol{z}_{s}~|~s \sim N(s\textbf{1},\sigma_{s}^{2}I_{100})$.
To get the classification boundary, we use logistic regression with regularization strength $\lambda$.
Recall that the bias degree of the classifier as $\beta_{clf}=||\boldsymbol{w}^*_s||/||\boldsymbol{w}^*_y||\in[0,+\infty)$. We set different data bias by using different ratios $n_{maj}:n_{min}$. We also set different standard deviations for $\boldsymbol{z}_y$ and $\boldsymbol{z}_s$.
All results were averaged over 100 random repetitions.

\noindent\textbf{Observations.}
As shown in Table~\ref{table-suppl}, in most cases, if we increase the regularization strength $\lambda$ in logistic regression, the classifier bias $\beta_{clf}$ will be larger. This observation motivates us to design a \textit{\textbf{simple}} but \textit{\textbf{effective}} method to obtain two different biased semantic directions in the latent space, that is to set different regularization strength $\lambda$.

\begin{table}[]
\setlength\tabcolsep{3.8pt}
\resizebox{0.47\textwidth}{!}{
\begin{tabular}{ccc|ccccc}
\hline
\multicolumn{3}{c|}{settings}                   & \multicolumn{5}{c}{regularization strength $\lambda$} \\
$n_{maj}:n_{min}$     & $\sigma_y$ & $\sigma_s$ & 1         & 10       & 100      & 1000     & 10000    \\ \hline
\multirow{4}{*}{2:1}  & 0.1        & 0.1        & 0.027     & 0.032    & 0.039    & 0.051    & 0.072    \\
                      & 0.1        & 1.0        & 0.027     & 0.032    & 0.040    & 0.051    & 0.072    \\
                      & 1.0        & 0.1        & 0.026     & 0.031    & 0.039    & 0.051    & 0.073    \\
                      & 1.0        & 1.0        & 0.030     & 0.033    & 0.040    & 0.051    & 0.073    \\ \hline
\multirow{4}{*}{3:1}  & 0.1        & 0.1        & 0.043     & 0.051    & 0.063    & 0.082    & 0.116    \\
                      & 0.1        & 1.0        & 0.043     & 0.051    & 0.063    & 0.082    & 0.116    \\
                      & 1.0        & 0.1        & 0.041     & 0.050    & 0.062    & 0.082    & 0.117    \\
                      & 1.0        & 1.0        & 0.044     & 0.051    & 0.063    & 0.082    & 0.117    \\ \hline
\multirow{4}{*}{4:1}  & 0.1        & 0.1        & 0.054     & 0.065    & 0.080    & 0.104    & 0.148    \\
                      & 0.1        & 1.0        & 0.052     & 0.063    & 0.079    & 0.104    & 0.148    \\
                      & 1.0        & 0.1        & 0.052     & 0.063    & 0.079    & 0.104    & 0.150    \\
                      & 1.0        & 1.0        & 0.055     & 0.064    & 0.079    & 0.104    & 0.149    \\ \hline
\multirow{4}{*}{5:1}  & 0.1        & 0.1        & 0.063     & 0.076    & 0.094    & 0.122    & 0.175    \\
                      & 0.1        & 1.0        & 0.063     & 0.075    & 0.093    & 0.122    & 0.174    \\
                      & 1.0        & 0.1        & 0.061     & 0.074    & 0.093    & 0.122    & 0.176    \\
                      & 1.0        & 1.0        & 0.064     & 0.075    & 0.093    & 0.122    & 0.175    \\ \hline
\multirow{4}{*}{6:1}  & 0.1        & 0.1        & 0.071     & 0.085    & 0.105    & 0.137    & 0.197    \\
                      & 0.1        & 1.0        & 0.070     & 0.084    & 0.104    & 0.137    & 0.195    \\
                      & 1.0        & 0.1        & 0.069     & 0.083    & 0.104    & 0.137    & 0.199    \\
                      & 1.0        & 1.0        & 0.071     & 0.084    & 0.104    & 0.136    & 0.197    \\ \hline
\multirow{4}{*}{7:1}  & 0.1        & 0.1        & 0.077     & 0.092    & 0.115    & 0.150    & 0.216    \\
                      & 0.1        & 1.0        & 0.077     & 0.092    & 0.114    & 0.149    & 0.214    \\
                      & 1.0        & 0.1        & 0.075     & 0.090    & 0.113    & 0.150    & 0.218    \\
                      & 1.0        & 1.0        & 0.077     & 0.091    & 0.113    & 0.149    & 0.216    \\ \hline
\multirow{4}{*}{8:1}  & 0.1        & 0.1        & 0.082     & 0.099    & 0.123    & 0.162    & 0.233    \\
                      & 0.1        & 1.0        & 0.082     & 0.099    & 0.122    & 0.160    & 0.231    \\
                      & 1.0        & 0.1        & 0.080     & 0.097    & 0.122    & 0.161    & 0.235    \\
                      & 1.0        & 1.0        & 0.082     & 0.097    & 0.121    & 0.160    & 0.233    \\ \hline
\multirow{4}{*}{9:1}  & 0.1        & 0.1        & 0.087     & 0.105    & 0.131    & 0.172    & 0.248    \\
                      & 0.1        & 1.0        & 0.087     & 0.105    & 0.130    & 0.171    & 0.246    \\
                      & 1.0        & 0.1        & 0.085     & 0.103    & 0.129    & 0.172    & 0.250    \\
                      & 1.0        & 1.0        & 0.087     & 0.103    & 0.129    & 0.171    & 0.248    \\ \hline
\multirow{4}{*}{10:1} & 0.1        & 0.1        & 0.092     & 0.110    & 0.138    & 0.181    & 0.262    \\
                      & 0.1        & 1.0        & 0.092     & 0.110    & 0.137    & 0.180    & 0.260    \\
                      & 1.0        & 0.1        & 0.089     & 0.108    & 0.136    & 0.181    & 0.264    \\
                      & 1.0        & 1.0        & 0.092     & 0.109    & 0.136    & 0.180    & 0.262    \\ \hline
\multirow{4}{*}{11:1} & 0.1        & 0.1        & 0.096     & 0.115    & 0.144    & 0.189    & 0.275    \\
                      & 0.1        & 1.0        & 0.096     & 0.115    & 0.143    & 0.188    & 0.272    \\
                      & 1.0        & 0.1        & 0.093     & 0.113    & 0.142    & 0.189    & 0.277    \\
                      & 1.0        & 1.0        & 0.096     & 0.114    & 0.142    & 0.188    & 0.275    \\ \hline
\multirow{4}{*}{12:1} & 0.1        & 0.1        & 0.100     & 0.120    & 0.150    & 0.197    & 0.286    \\
                      & 0.1        & 1.0        & 0.099     & 0.119    & 0.149    & 0.196    & 0.284    \\
                      & 1.0        & 0.1        & 0.097     & 0.118    & 0.148    & 0.197    & 0.289    \\
                      & 1.0        & 1.0        & 0.099     & 0.118    & 0.148    & 0.196    & 0.287    \\ \hline
\multirow{4}{*}{13:1} & 0.1        & 0.1        & 0.103     & 0.124    & 0.155    & 0.205    & 0.298    \\
                      & 0.1        & 1.0        & 0.103     & 0.124    & 0.154    & 0.203    & 0.294    \\
                      & 1.0        & 0.1        & 0.101     & 0.122    & 0.154    & 0.205    & 0.301    \\
                      & 1.0        & 1.0        & 0.103     & 0.123    & 0.153    & 0.203    & 0.298    \\ \hline
\multirow{4}{*}{14:1} & 0.1        & 0.1        & 0.107     & 0.128    & 0.160    & 0.211    & 0.308    \\
                      & 0.1        & 1.0        & 0.106     & 0.128    & 159      & 0.210    & 0.306    \\
                      & 1.0        & 0.1        & 0.104     & 0.126    & 0.159    & 0.212    & 0.311    \\
                      & 1.0        & 1.0        & 0.106     & 0.127    & 0.158    & 0.210    & 0.309    \\ \hline
\end{tabular}}
\caption{Results of classifier bias $\beta_{clf}$ on synthetic dataset. Empirically, in most cases, the classifier bias $\beta_{clf}$ will be larger, if we increase the regularization strength $\lambda$ in logistic regression.}
\label{table-suppl}
\end{table}

\section*{Appendix C. Additional results on real dataset}
\begin{figure*}
\begin{center}
\includegraphics[width=\linewidth]{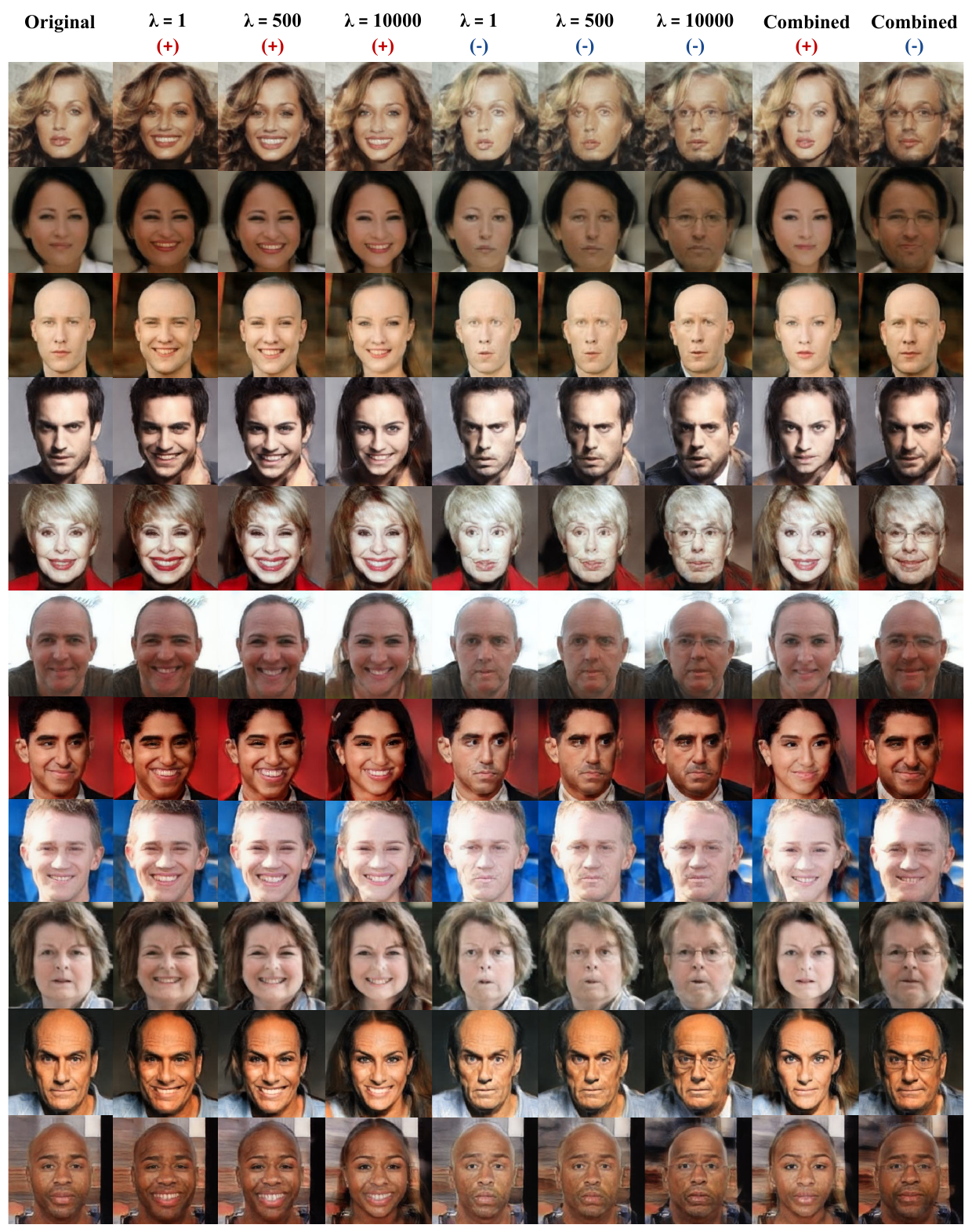}
\end{center}
\vspace{-1.5em}
\caption{Illustration of images edited by different semantic directions, which are trained with different regularization strength $\lambda$.}
\label{fig-suppl}
\end{figure*}

\vspace{-0.5em}
In response to the above findings, we show the images edited by different semantic directions, obtained with different regularization strengths $\lambda$.
The training dataset (sampled from CelebA) is biased where the target attribute $Smiling$ is spuriously correlated with the spurious attributes $Female$ and $Young$.
We first use a trained generative model to encode the images into latent codes. Then we train linear classifiers in latent space using logistic regression with different $\lambda$. The semantic directions are normal vectors of the learned classification boundaries.
As shown in Figure~\ref{fig-suppl}, a larger $\lambda$ produces a larger bias in direction, resulting in a more obvious change in spurious attributes.

\section*{Appendix D. Implementation Details}
For generative modeling, we utilize StyleGAN2~\cite{karras2020analyzing} for generator and e4e~\cite{tov2021designing} for encoder. We use HFGI~\cite{wang2021HFGI} algorithm to train generative models on training dataset with image size of 256 for 30 epochs.
The size of features encoded by e4e is (18, 512), and we average over the channels to get latent codes with size of 512. We use regularized logistic regression to obtain directions, and the values of regularization strength $\lambda$ are 1e+4 and 1e-4 respectively. To get the optimal combination coefficients, we perform grid search and use CLIP~\cite{radford2021learning} as a reference model. More details about combination coefficients are shown in the next subsection. For representation model, we use ResNet-18~\cite{he2016deep} for encoder and the representation dimensions are 512. We train the encoder for 135 epochs. We use Adam~\cite{kingma2014adam} as optimizer with learning rate 3e-4. We set the editing range [$\alpha_l$,$\alpha_u$] as [3,5]. For efficiency, we approximate the sampled degree as an integer. To complete the classification, we fix the encoder and train a linear classifier with Adam until convergence. The learning rate is 1e-2 with 1e-6 weight decay.


\end{document}